\documentclass[11pt]{article}

\usepackage[preprint]{acl}

\usepackage{times}
\usepackage{latexsym}
\usepackage{booktabs}
\usepackage{amsmath}
\usepackage{amssymb}
\usepackage{multirow}
\usepackage{xcolor}
\usepackage{url}
\usepackage{array} 
\usepackage{algorithm}
\usepackage{algorithmic}
\usepackage{hyperref}

\usepackage[T1]{fontenc}

\usepackage[utf8]{inputenc}

\usepackage{microtype}

\usepackage{inconsolata}

\usepackage{graphicx}

\newcommand{\method}{\textsc{IoU-PD}}

\title{\method{}: IoU-Aware Privileged Distillation for Visual Grounding with Multimodal Large Language Models}

\author{
 \textbf{Xiuyuan Zhu\textsuperscript{1,2}},
 \textbf{Ke Lu\textsuperscript{1,3}},
 \textbf{Hao Wu\textsuperscript{1,2}},
 \textbf{Siwen Jiao\textsuperscript{4}},
\\
 \textbf{Zijin Du\textsuperscript{1}},
 \textbf{Dongming Zhang\textsuperscript{2}},
 \textbf{Jian Xue\textsuperscript{1,*}}
\\
\\
 \textsuperscript{1}University of Chinese Academy of Sciences, Beijing, China
\\
 \textsuperscript{2}State Key Laboratory of Communication Content Cognition, Beijing, China
\\
 \textsuperscript{3}Peng Cheng Laboratory, Shenzhen, Guangdong, China
\\
 \textsuperscript{4}National University of Singapore, Singapore
\\
 \small{
   \textsuperscript{*}Correspondence: xuejian@ucas.ac.cn
 }
}

\begin{document}
\maketitle
\begin{abstract}
Visual grounding with multimodal large language models is commonly formulated as autoregressive coordinate generation, where a model outputs bounding-box coordinates as text given an image and a referring-expression prompt. While this interface is simple and compatible with instruction following, it introduces a mismatch between training and evaluation: training optimizes token-level likelihood over coordinate strings, whereas grounding quality is measured by geometric overlap. We propose \method{}, an IoU-aware privileged distillation method for coordinate-generating multimodal large language models. \method{} uses ground-truth boxes not only as coordinate targets, but also as privileged training-time guidance. During training, the student receives the original image and prompt, while a frozen teacher receives a box-marked image and an augmented prompt that indicates the marked region. The student is trained with a supervised fine-tuning anchor and a privileged distillation loss whose token weights reflect both geometric importance and teacher reliability. At inference time, \method{} requires no box overlay, privileged hint, teacher branch, or additional prediction module. Experiments on standard referring-expression grounding benchmarks show consistent region-level improvements over strong coordinate-generating baselines, demonstrating that ground-truth boxes can provide useful privileged guidance beyond serving as coordinate labels. 
\href{https://xyzzzh.github.io/IoU-PD/}{\textcolor{blue}{Project page}}.
\end{abstract}

\section{Introduction}
\label{sec:intro}

Visual grounding is a fundamental task in vision-and-language understanding: given an image and a natural language expression, a model localizes the corresponding region. This ability supports multimodal reasoning, visual question answering, embodied agents, and human-computer interaction by connecting language outputs to visual content.

Recent multimodal large language models provide a general grounding interface through coordinate generation. The model receives an image and a referring-expression prompt, then generates a structured textual response containing bounding-box coordinates instead of using a task-specific localization head. This keeps grounding within the same autoregressive framework used for instruction following and other multimodal tasks.

However, coordinate generation introduces a mismatch between training and evaluation. Supervised fine-tuning optimizes token-level losses over coordinate strings, whereas grounding is evaluated by geometric overlap between predicted and ground-truth boxes. A small token change may cause a large IoU difference, while geometrically similar boxes may have different token sequences. Coordinate imitation therefore does not explicitly align learning with the geometric structure of grounding.

This mismatch suggests that ground-truth boxes can provide more supervision than coordinate labels alone. In standard grounding datasets, the ground-truth box is typically used only as the target answer. During training, however, the same box also identifies the visual region referred to by the expression. Although this information is unavailable at inference time, it can serve as privileged training-time guidance.

We propose \method{}, an IoU-aware privileged distillation method for coordinate-generating multimodal large language models. During training, the student receives the original image and referring-expression prompt, while a frozen teacher receives a box-marked image and an augmented prompt indicating the marked region. The teacher thus receives explicit regional guidance, while the student preserves the standard inference-time input format. At inference, \method{} requires no box overlay, privileged hint, teacher branch, or additional prediction module.

The teacher and student condition on different inputs, so the teacher distribution differs from the student's inference-time distribution. We retain supervised fine-tuning as an anchor: it ties the student to ground-truth coordinates, while privileged distillation adds region-aware guidance.

We further adapt distillation to coordinate outputs. Their tokens encode box boundaries, and different coordinates or digit positions can affect IoU differently. \method{} weights token-level distillation using geometry and reliability cues, aligning the training signal with region-level grounding quality.

Experiments on benchmarks show that \method{} improves pooled mIoU by 1.08 points and Acc@0.7 by 1.44 points over SFT. Ordinary-teacher and privileged-teacher controls separate the gains from privileged input and geometry-aware weighting. These results show that existing box annotations provide useful supervision beyond coordinate labels while preserving the student's inference interface.

The contributions of this paper are as follows.

\begin{itemize}
\item We propose \method{}, which combines supervised coordinate learning with privileged distillation from a frozen teacher conditioned on a box-marked image and hint. The student retains its standard inference interface.
\item We design geometry-aware distillation weights using sample IoU, coordinate error, and teacher--student preference, together with digit importance and teacher confidence.
\item We conduct experiments and ablations on benchmarks, showing consistent region-level improvements and clarifying the roles of SFT, privileged teacher input, and IoU-aware weighting.
\end{itemize}

\begin{figure*}[t]
\centering
\includegraphics[width=\textwidth,trim=0 10 0 0,clip]{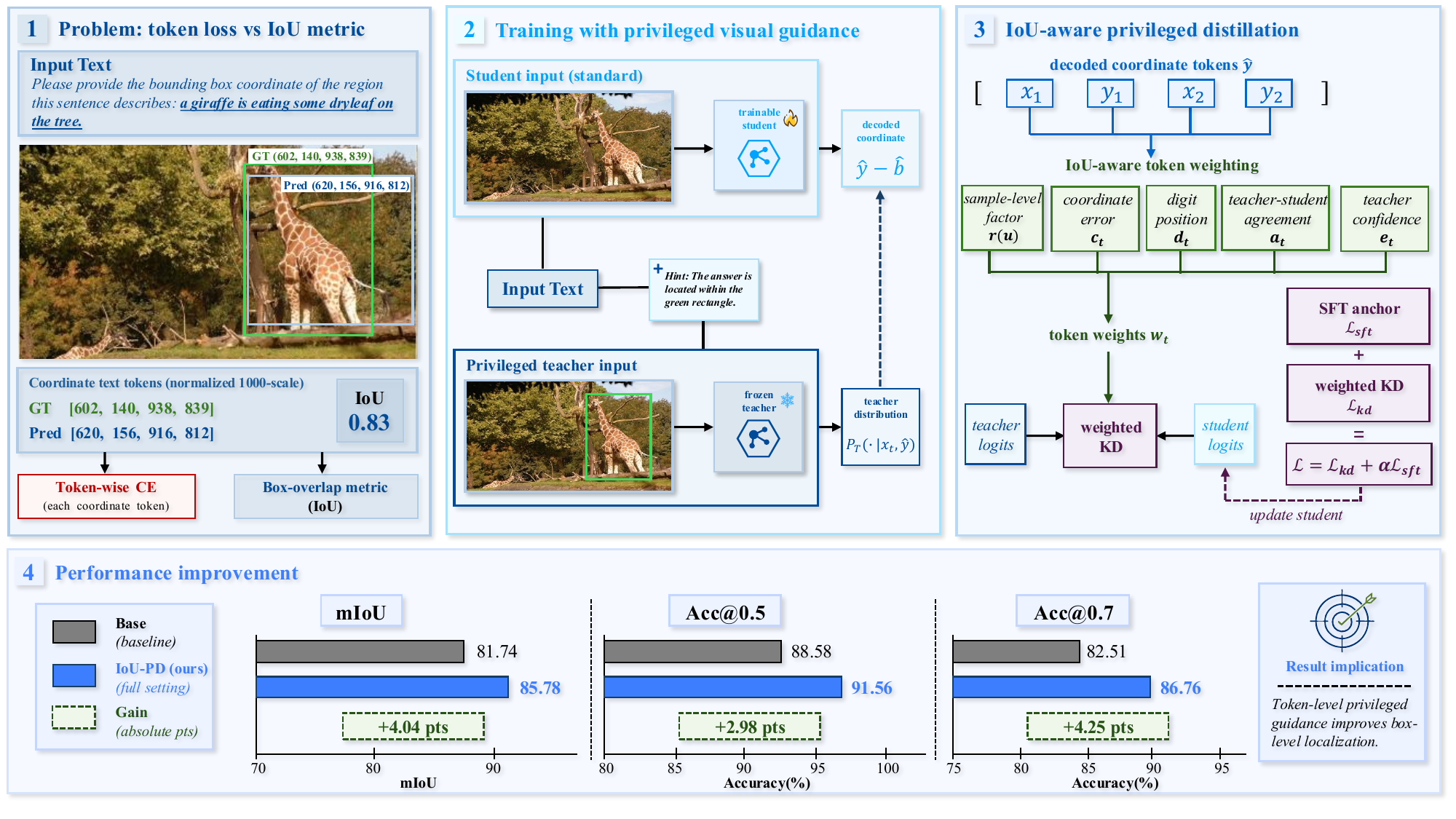}
\caption{
Overview of \method{}.
Ground-truth boxes are used not only as coordinate targets, but also to construct privileged teacher inputs during training.
The student receives the original image and original referring-expression prompt, while the teacher receives a box-marked image and an augmented prompt that indicates the marked region.
The training objective combines an SFT anchor with IoU-aware privileged distillation, while keeping the inference-time input format unchanged.
}
\label{fig:framework}
\end{figure*}

\section{Related Work}
\label{sec:related}

\subsection{Visual Grounding}

Visual grounding, or referring expression comprehension, localizes the image region described by text. Common benchmarks include RefCOCO, RefCOCO+, and RefCOCOg~\cite{kazemzadeh-etal-2014-referitgame,refcoco,refcocog,abc}. Detectors such as DETR~\cite{detr}, Grounding DINO~\cite{grounding-dino}, OWLv2~\cite{owlv2}, and YOLO-World~\cite{yoloworld} predict boxes with localization heads. We focus on generative grounding, which represents boxes as coordinate sequences.

\subsection{Multimodal Large Language Models for Grounding}

Multimodal large language models combine visual understanding with language generation~\cite{shikra,kosmos,qwenvl,qwen25vl,qwen3vl,llava1,llava2,llava3,deepseekvl2,glm45,glm41,glm5,gpt5,gemini,seed15,seed18,seed2,internvl1,internvl2,internvl3,internvl4}. This interface supports grounding through autoregressive coordinate generation. Specialist methods improve localization through grounding-specific post-training, including Visual-RFT~\cite{virft}, VLM-R1~\cite{vlm-r1}, Rex-Omni~\cite{rex-omni}, Smooth Operator~\cite{smoothoperator}, Hi-Token~\cite{zhu2026hitoken} and DeepGrounder~\cite{zhang2026deepgrounder}.

Coordinate generation represents boxes as token sequences, whereas grounding is evaluated by geometric overlap. \method{} aligns token-level distillation with box geometry through privileged teacher guidance and IoU-aware weighting, while retaining the student's coordinate-generation interface at inference.

\subsection{Knowledge Distillation and Privileged Information}

Knowledge distillation transfers information from a teacher distribution to a student distribution~\cite{hinton2015distilling}. Learning using privileged information studies a related setting in which additional information is available during training but unavailable at inference time~\cite{vapnik2009privileged,opcd}. More recently, on-policy distillation has been explored across language and multimodal models, allowing the teacher and student to condition on different information while supervising trajectories sampled from the student policy~\cite{zhao2026opsd,shortopd,vision}. Our work follows this general paradigm but introduces multimodal privileged information tailored to visual grounding: the teacher receives a box-marked image and an augmented text prompt, whereas the deployed student receives only the original image and the original referring-expression prompt.

\subsection{Structured Supervision for Coordinate Outputs}

Coordinate strings encode box boundaries, and their digit positions have different effects on geometric overlap. GUI-SD~\cite{zhang2026guisd} uses a visually privileged self-teacher to supervise student rollouts, with token weights based on digit significance and teacher entropy. \method{} builds on privileged teacher input and coordinate-token weighting for referring-expression grounding. It further uses predicted-box IoU to weight samples, coordinate errors to emphasize inaccurate box boundaries, and a teacher--student preference gate to modulate token supervision. These factors connect the allocation of distillation loss to box geometry and teacher guidance.

\section{Method}
\label{sec:method}

\subsection{Overview}

The goal is to improve coordinate-generating multimodal large language models for visual grounding without changing the inference-time input format. Standard supervised fine-tuning uses the ground-truth box only as the target coordinate answer. \method{} uses the same box in an additional way: it is drawn on the image to construct a privileged visual input for a teacher model during training.

The student receives the original image and original referring-expression prompt.
The teacher receives the box-marked image and an augmented text prompt, where the original prompt is followed by a privileged hint:
``The answer is located within the green rectangle.''
This privileged hint is used only during training and is never provided to the student or used at inference time.
The teacher has the same architecture as the student and is initialized from the same base checkpoint, but it is kept frozen during training. Its output distribution is detached and used as a stop-gradient target. No EMA teacher is used. Training combines supervised fine-tuning on the ground-truth coordinate string with privileged teacher distillation on student-generated response tokens. The distillation loss is further weighted by geometry and reliability cues.

\begin{algorithm}[t]
\caption{Training step of \method{}}
\label{alg:method}
\begin{algorithmic}[1]
\REQUIRE Minibatch $\{(I_i, q_i, b_i^\ast)\}_{i=1}^{B}$, student $p_\theta$, frozen teacher $p_T$, SFT weight $\alpha$
\ENSURE Updated student $p_\theta$

\FOR{$i = 1$ to $B$}
  \STATE $x_{s,i} \leftarrow (I_i, q_i)$
  \STATE $x_{t,i} \leftarrow (\mathrm{DrawBox}(I_i,b_i^\ast), q_i \oplus h)$
  \STATE $y_i^\ast \leftarrow \mathrm{Format}(b_i^\ast)$
  \STATE $\hat{y}_i \leftarrow \mathrm{Decode}(p_\theta(\cdot \mid x_{s,i}))$
  \STATE Treat $\hat{y}_i$ as a fixed sequence for distillation
  \STATE Score $y_i^\ast$ under $x_{s,i}$ for supervised fine-tuning
  \STATE Score $\hat{y}_i$ under $x_{s,i}$ and $x_{t,i}$ for distillation
  \STATE Parse $\hat{y}_i$ into $\hat{b}_i$ and compute raw weights $w_{i,t}$
  \STATE Detach $w_{i,t}$ and retain valid response positions $\mathcal{Y}_i$
\ENDFOR

\STATE Form $\mathcal{V}=\{(i,t):t\in\mathcal{Y}_i\}$ and $N=|\mathcal{V}|$
\STATE $\mu_w\leftarrow N^{-1}\sum_{(i,t)\in\mathcal{V}}w_{i,t}$
\STATE $\bar{w}_{i,t}\leftarrow w_{i,t}/\max(\mu_w,10^{-6})$
\STATE Average weighted token KL terms over $\mathcal{V}$ to obtain $\mathcal{L}_{kd}$
\STATE Compute $\mathcal{L}_{sft}$
\STATE $\mathcal{L} \leftarrow \mathcal{L}_{kd} + \alpha \mathcal{L}_{sft}$
\STATE Update $p_\theta$ with $\mathcal{L}$

\end{algorithmic}
\end{algorithm}

\subsection{Task Formulation and Privileged Teacher}

Each training example is denoted as
\[
    (I, q, b^\ast),
\]
where \(I\) is the image, \(q\) is the referring expression, and
\[
    b^\ast = (x_1^\ast, y_1^\ast, x_2^\ast, y_2^\ast)
\]
is the ground-truth box. The target coordinate response is
\[
    y^\ast = \mathrm{Format}(b^\ast).
\]
The student input is
\[
    x_s = (I,q),
\]
where \(q\) is the original referring-expression prompt. During training, we construct a privileged teacher input
\[
    x_t = (I^{box}, q^+),
\]
where \(I^{box}\) is the original image with the ground-truth box marked in green, and
\[
    q^+ = q \oplus h.
\]
Here, \(h\) is a short teacher-side hint appended to the original prompt:
\begin{quote}
\emph{Hint: The answer is located within the green rectangle.}
\end{quote}

The image content outside the box is preserved, so the teacher still sees the full scene context. The hint does not reveal the coordinate values, but it aligns the teacher's attention with the marked region. Both \(I^{box}\) and \(h\) are used only for the frozen teacher during training.

The teacher is frozen throughout training. Its distribution is computed under \(x_t\), detached from the computation graph, and used only as a training-time target. Gradients are propagated only through the student model.

\subsection{Training Objective}

The SFT loss is computed on the ground-truth response:
\[
    \mathcal{L}_{sft}
    =
    - \sum_{t \in \mathcal{Y}^\ast}
    \log p_\theta(y_t^\ast \mid x_s, y_{<t}^\ast),
\]
where \(\mathcal{Y}^\ast\) denotes response positions in \(y^\ast\).

The distillation loss is computed on a response sequence decoded from the current student:
\[
    \hat{y} \sim p_\theta(\cdot \mid x_s).
\]

The decoded sequence is treated as fixed, so gradients are not back-propagated through the discrete decoding step. For a minibatch of \(B\) examples, let \(\mathcal{Y}_i\) contain the valid response positions of example \(i\). Define
\[
    \mathcal{V}=\{(i,t):1\leq i\leq B,\ t\in\mathcal{Y}_i\},
    \qquad N=|\mathcal{V}|.
\]
Prompt and padding positions are excluded. The privileged distillation loss is averaged over these \(N\) tokens:
\[
\begin{aligned}
\mathcal{L}_{kd}
&=
\frac{1}{N}\sum_{(i,t) \in \mathcal{V}}
\bar{w}_{i,t}\,
D_{\mathrm{KL}}
\Big(
\mathrm{sg}\!\left[
p_T(\cdot \mid x_{t,i},\hat{y}_{i,<t})
\right]
\\
&\qquad\qquad\qquad
\Vert\,
p_\theta(\cdot \mid x_{s,i},\hat{y}_{i,<t})
\Big).
\end{aligned}
\]
Here, \(\mathrm{sg}[\cdot]\) denotes stop-gradient. The token weights \(\bar{w}_{i,t}\) share a normalization across \(\mathcal{V}\), as defined below. The total objective is

\[
    \mathcal{L}
    =
    \mathcal{L}_{kd}
    +
    \alpha \mathcal{L}_{sft}.
\]

\subsection{IoU-Aware Token Weighting}

The decoded response \(\hat{y}\) is parsed into a predicted box
\[
    \hat{b} = (\hat{x}_1,\hat{y}_1,\hat{x}_2,\hat{y}_2)
\]
when possible. The localization quality is
\[
    u = \mathrm{IoU}(\hat{b}, b^\ast).
\]
The parser also maps coordinate digit tokens to their coordinate identity
\[
    m(t) \in \{1,2,3,4\}
\]
and digit position \(\rho(t)\). The digit position is defined by decimal significance, with larger \(\rho(t)\) assigned to more significant digits. Non-coordinate tokens, including brackets, commas, spaces, separators, and punctuation, are assigned neutral geometry weights.

For coordinate \(k\), the coordinate error is
\[
    \delta_k = |\hat{b}_k - b_k^\ast|.
\]
The unnormalized token weight is
\[
    w_t
    =
    r(u)
    \cdot
    c_t
    \cdot
    d_t
    \cdot
    a_t
    \cdot
    e_t.
\]
The sample-level factor is
\[
    r(u) = \exp\left(\frac{1-u}{\tau_r}\right).
\]
For a token \(t\) belonging to coordinate \(m(t)\),
\[
    c_t =
    \frac{
        \exp(\delta_{m(t)} / \tau_c)
    }{
        \frac{1}{4}\sum_{k=1}^{4}\exp(\delta_k / \tau_c)
    },
    \qquad
    d_t = 1 + \lambda_d \rho(t).
\]
For non-coordinate tokens, \(c_t=d_t=1\).

The agreement and confidence factors are
\[
\begin{aligned}
a_t
&=
\sigma\Big(
\beta
\big[
\log p_T(\hat{y}_t \mid x_t,\hat{y}_{<t})  \\
&\qquad\qquad
-
\log p_\theta(\hat{y}_t \mid x_s,\hat{y}_{<t})
\big]
\Big).
\end{aligned}
\]
and
\[
    e_t =
    \exp\left(
        - \frac{
        H(p_T(\cdot \mid x_t,\hat{y}_{<t}))
        }{\tau_e}
    \right).
\]
Writing \(w_{i,t}\) for the raw weight of token \(t\) in example \(i\), we normalize jointly over the valid response tokens in the minibatch:
\[
\begin{aligned}
    \mu_w &= \frac{1}{N}\sum_{(j,s)\in\mathcal{V}}\mathrm{sg}[w_{j,s}], \\
    \bar{w}_{i,t} &= \frac{\mathrm{sg}[w_{i,t}]}{\max(\mu_w,\epsilon)},
    \qquad \epsilon=10^{-6}.
\end{aligned}
\]
This shared normalization preserves the relative weighting of responses with different IoU factors. The weights are detached, so gradients flow only through the student distribution in the distillation loss.

\begin{figure}[t]
\centering
\includegraphics[width=\columnwidth, trim=0 35 0 0, clip]{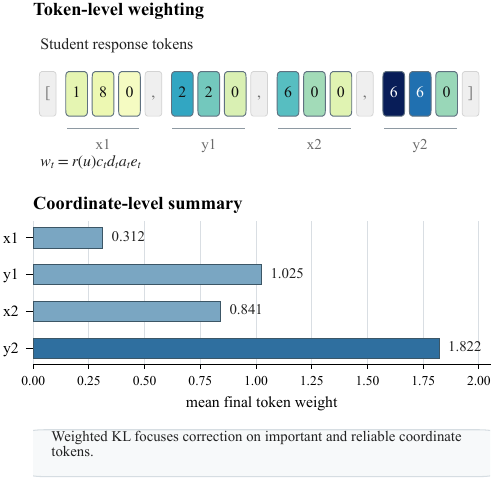}
\caption{
Visualization of IoU-aware token weighting.
\method{} assigns larger distillation weights to coordinate tokens that are more geometrically influential or supported by more reliable teacher guidance.
}
\label{fig:token_weighting}
\end{figure}

Figure~\ref{fig:token_weighting} illustrates the effect of the proposed weighting strategy.
Rather than treating all response tokens uniformly, \method{} emphasizes coordinate tokens associated with larger geometric errors, more significant digit positions, and more reliable teacher guidance.

If parsing fails, the geometry-dependent factors fall back to neutral values:
\[
    r(u)=1,\qquad c_t=1,\qquad d_t=1.
\]
This keeps the distillation loss well defined for malformed responses.

\subsection{Training and Inference}

During training, the ground-truth box serves both as the coordinate target and as the source of the privileged teacher input. During inference, the teacher is removed and the student follows the standard input-output format:
\[
    (I,q) \rightarrow \hat{y}.
\]
Thus, \method{} requires no ground-truth box, box overlay, EMA teacher, or additional prediction module at deployment time.

\section{Experiments}
\label{sec:experiments}

\subsection{Experimental Setup}

\paragraph{Datasets}

Training uses RefCOCO-style grounding examples with an image, a referring expression, and a normalized ground-truth box. Evaluation is conducted on five held-out splits: RefCOCO testA, RefCOCO testB, RefCOCOg test, RefCOCO+ testA, and RefCOCO+ testB. 

\paragraph{Main Setting}

Unless otherwise specified, the main \method{} setting uses a Qwen3-VL-4B backbone, 300k training examples, and 3 training epochs. The student receives the original image and referring expression, while the frozen teacher receives the same expression and a box-marked image. The model is trained with supervised fine-tuning, privileged teacher distillation, and IoU-aware token weighting. Reduced model and data settings are used only for ablation studies.

\paragraph{Evaluation Protocol}

All comparison models in Table~\ref{tab:sota_comparison} are evaluated under the same prompt, coordinate parser, coordinate normalization, box canonicalization, and metric computation script. This unified protocol avoids comparing results produced by different prompting or parsing rules. The main metrics are mIoU, Acc@0.5, and Acc@0.7.

\subsection{Main Results}

\begin{table*}[t]
\centering
\setlength{\tabcolsep}{5pt}
\renewcommand{\arraystretch}{1.05}
\begin{tabular*}{\textwidth}{@{\extracolsep{\fill}}lcccccc@{}}
\toprule
\multirow{2}{*}{Method}
& \multicolumn{2}{c}{RefCOCO}
& \multicolumn{2}{c}{RefCOCO+}
& \multicolumn{2}{c}{RefCOCOg} \\
\cmidrule(lr){2-3}
\cmidrule(lr){4-5}
\cmidrule(lr){6-7}
& mIoU & Acc@0.5
& mIoU & Acc@0.5
& mIoU & Acc@0.5 \\
\midrule
\multicolumn{7}{@{}l}{\textit{Open-set detection models}} \\
OWLv2
& 41.5 & 40.2
& 37.3 & 35.1
& 30.2 & 29.2 \\
Grounding DINO
& 56.2 & 57.5
& 56.7 & 57.2
& 58.8 & 59.8 \\
\midrule
\multicolumn{7}{@{}l}{\textit{Open-source VLMs}} \\
Qwen2.5-VL-3B
& 55.6 & 60.2
& 52.6 & 62.3
& 49.8 & 44.1 \\
Qwen2.5-VL-7B
& 60.7 & 67.7
& 58.2 & 64.8
& 50.6 & 53.4 \\
Qwen2.5-VL-72B
& 62.5 & 70.4
& 58.9 & 66.1
& 55.1 & 59.7 \\
DeepSeek-VL2
& 51.1 & 56.2
& 44.6 & 46.6
& 38.8 & 34.2 \\
GLM-4.1V-9B
& 83.5 & 91.6
& 80.2 & 87.6
& 80.1 & 83.6 \\
GLM-4.6V-106B
& 82.0 & 88.5
& 75.6 & 80.9
& 80.2 & 86.2 \\
Qwen3-VL-8B
& 84.15 & 91.28
& 80.5 & 86.98
& 82.6 & 89.5 \\
\midrule
\multicolumn{7}{@{}l}{\textit{Specialist VLMs}} \\
VLM-R1
& 63.1 & 69.8
& 64.4 & 71.5
& 66.7 & 73.4 \\
Rex-Omni
& 81.9 & 88.2
& 77.5 & 83.4
& 77.3 & 86.1 \\
\method{} (Ours)
& \textbf{86.44} & \textbf{93.18}
& \textbf{83.80} & \textbf{90.00}
& \textbf{87.23} & \textbf{91.45} \\
\bottomrule
\end{tabular*}
\caption{
Comparison with existing models on RefCOCO, RefCOCO+, and RefCOCOg. All baselines are re-evaluated under the same prompt, parser, coordinate normalization, and metric computation protocol. All values are reported as percentages. The best result in each column is shown in bold.
}
\label{tab:sota_comparison}
\end{table*}

Table~\ref{tab:sota_comparison} compares \method{} with open-set detection models, open-source VLMs, and specialist VLMs under the same evaluation protocol. The comparison focuses on mIoU and Acc@0.5, which measure region-level localization quality. Under the main setting, \method{} achieves the best results on all reported datasets and metrics.

The comparison with general open-source VLMs shows that stronger base multimodal models can already provide competitive coordinate-generation grounding performance. However, \method{} further improves this behavior by using the ground-truth box as training-time privileged visual information. Compared with specialist VLMs, \method{} also remains competitive or stronger under the same evaluation protocol. These results indicate that ground-truth boxes can serve a broader training role than coordinate supervision alone by providing visual guidance to the teacher.

\begin{figure}[t]
\centering
\includegraphics[width=\columnwidth]{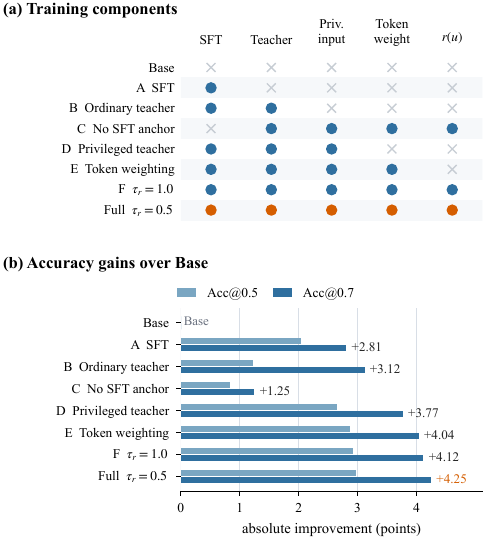}
\caption{
Component ablations of \method{}.
The upper panel shows the enabled training components; Priv. input denotes the box-marked image and fixed hint.
The lower panel reports Acc@0.5 and Acc@0.7 gains over Base under the main setting.
}
\label{fig:component_ablations_single}
\end{figure}

\subsection{Ablation Studies}

\paragraph{Component Ablations}

\begin{table}[t]
\centering
\setlength{\tabcolsep}{3pt}
\renewcommand{\arraystretch}{1.05}
\begin{tabular*}{\columnwidth}{@{\extracolsep{\fill}}llccc@{}}
\toprule
Added component & mIoU & A@0.5 & A@0.7 \\
\midrule
Base & 0.8174 & 88.58 & 82.51 \\
SFT & 0.8470 & 90.62 & 85.32 \\
Original teacher & 0.8492 & 89.80 & 85.63 \\
Box teacher & 0.8543 & 91.23 & 86.28 \\
Token weighting & 0.8565 & 91.45 & 86.55 \\
Full \method{} & \textbf{0.8578} & \textbf{91.56} & \textbf{86.76} \\
\bottomrule
\end{tabular*}
\caption{
Component ablations under the main setting. The rows show the effect of progressively adding supervised fine-tuning, teacher distillation, privileged box input, and IoU-aware token weighting.
}
\label{tab:component_ablations_summary}
\end{table}

\begin{figure*}[!hbtp]
\centering
\includegraphics[width=\textwidth]{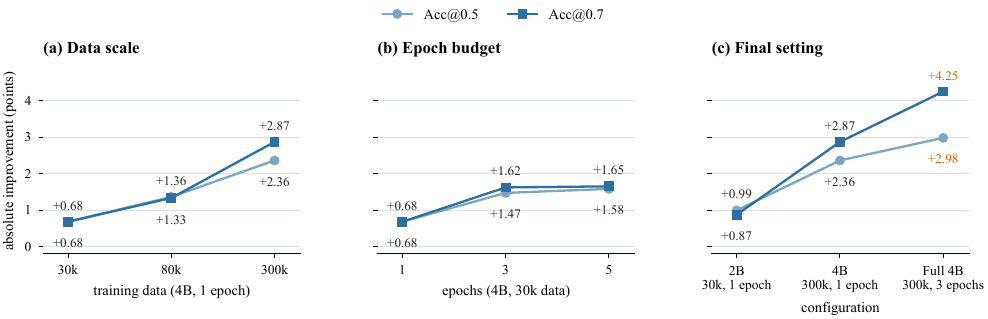}
\caption{
Scaling ablations under different data sizes, epoch budgets, and final model settings. The curves show absolute Acc@0.5 and Acc@0.7 improvements over the corresponding same-size base model.
}
\label{fig:scaling_ablations}
\end{figure*}

Table~\ref{tab:component_ablations_summary} summarizes the main component ablations, and Figure~\ref{fig:component_ablations_single} provides the corresponding component matrix and Acc@.7 gains over the base model. SFT gives the largest single improvement, increasing mIoU from 0.8174 to 0.8470 and Acc@.7 from 82.51 to 85.32. This confirms that direct coordinate supervision is the main anchor for learning the output format and the grounding distribution.

The original-teacher variant separates self-distillation from privileged visual guidance. In this setting, the teacher receives the same original image as the student. It improves Acc@.7 from 85.32 to 85.63 over SFT, but the gain is smaller than using a box-marked teacher. Replacing the original teacher with the privileged box teacher improves mIoU to 0.8543 and Acc@.7 to 86.28. This comparison shows that the improvement is not only due to distillation itself; the box-marked teacher input provides additional useful visual guidance.

IoU-aware token weighting further improves the privileged teacher setting. Adding token weighting raises mIoU from 0.8543 to 0.8565 and Acc@.7 from 86.28 to 86.55. The full objective obtains the best result, with 0.8578 mIoU and 86.76 Acc@.7. The full matrix in Figure~\ref{fig:component_ablations_single} also shows that removing the SFT anchor leads to a much smaller gain, while changing the sample-level IoU factor temperature remains close to the full model. Overall, the ablations support the main design: SFT provides the anchor, the privileged box teacher supplies training-time visual guidance, and IoU-aware weighting refines the token-level distillation signal.

\begin{figure}[!hbtp]
\centering
\includegraphics[width=\columnwidth]{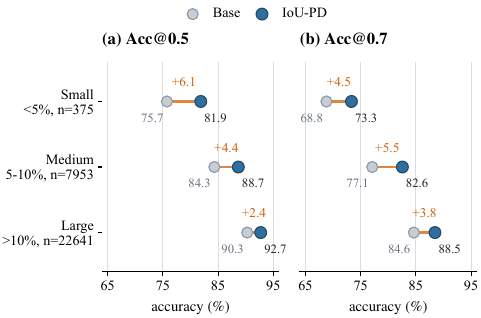}
\caption{
\method{} improves grounding across object sizes.
Examples are grouped by the ground-truth bounding-box area in the normalized coordinate space: small (\(<5\%\)), medium (\(5\%-10\%\)), and large (\(>10\%\)).
Points report Acc@0.5 and Acc0.7 for the base model and the \method{} model, with orange segments and labels indicating absolute gains.
}
\label{fig:object_size_performance}
\end{figure}

\paragraph{Scaling Ablations}

\begin{table}[t]
\centering
\setlength{\tabcolsep}{2.2pt}
\renewcommand{\arraystretch}{1.05}
\begin{tabular*}{\columnwidth}{@{\extracolsep{\fill}}lccrrrr@{}}
\toprule
Model & Data & Ep. & mIoU & A@.5 & A@.7 & \(\Delta\) \\
\midrule
Base 2B & -- & -- & 0.793 & 85.64 & 79.76 & -- \\
2B & 30k & 1 & 0.799 & 86.64 & 80.63 & +0.87 \\
\midrule
Base 4B & -- & -- & 0.817 & 88.58 & 82.51 & -- \\
4B & 30k & 1 & 0.823 & 89.26 & 83.20 & +0.68 \\
4B & 30k & 3 & 0.830 & 90.05 & 84.13 & +1.62 \\
4B & 30k & 5 & 0.831 & 90.16 & 84.16 & +1.65 \\
4B & 80k & 1 & 0.828 & 89.94 & 83.84 & +1.33 \\
4B & 300k & 1 & 0.839 & 90.94 & 85.38 & +2.87 \\
\method{} & 300k & 3 & \textbf{0.857} & \textbf{91.56} & \textbf{86.76} & \textbf{+4.25} \\
\bottomrule
\end{tabular*}
\caption{
Scaling ablations under different model sizes, data sizes, and training budgets. \(A@.5\) and \(A@.7\) denote Acc@0.5 and Acc@0.7, and \(\Delta\) denotes the Acc@0.7 improvement over the same-size base model.
}
\label{tab:scaling_ablations}
\end{table}

Table~\ref{tab:scaling_ablations} and Figure~\ref{fig:scaling_ablations} summarize how \method{} behaves under different backbone sizes, data scales, and optimization budgets. The reduced settings show that the method remains effective with a smaller 2B backbone and with limited training data. Figure~\ref{fig:scaling_ablations}(a) shows that, under the 4B backbone and a fixed one-epoch budget, increasing the training data from 30k to 80k and then to 300k leads to progressively larger gains, with the 300k setting giving the largest improvement. Figure~\ref{fig:scaling_ablations}(b) shows that, under the 4B backbone and 30k training data, increasing the training budget from 1 to 3 epochs improves both Acc@0.5 and Acc@0.7, while the gain from 3 to 5 epochs is marginal. Figure~\ref{fig:scaling_ablations}(c) compares representative reduced settings with the final configuration and shows that the final configuration achieves the largest gain overall. These results indicate that privileged visual teacher guidance remains useful across scales and benefits from both larger grounding data and sufficient optimization budget.

\subsection{Object-Size Analysis}

Figure~\ref{fig:object_size_performance} compares the base model and the main \method{} under different ground-truth object sizes.
\method{} improves both Acc@0.5 and Acc@0.7 in all three size groups.
The gains are especially clear for small and medium objects, where coordinate errors occupy a larger fraction of the target region and stricter overlap thresholds are harder to satisfy.
This suggests that privileged box guidance is not only improving easy large-object cases, but also helps the model localize more size-sensitive targets.

\begin{figure}[!htbp]
\centering
\includegraphics[width=\columnwidth]{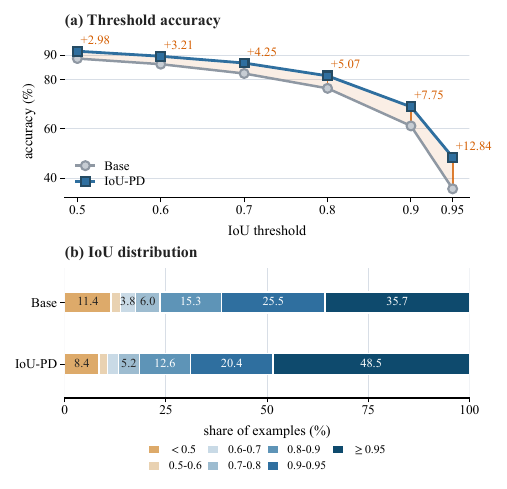}
\caption{
Threshold-sensitivity analysis. \method{} improves precision under stricter IoU thresholds and shifts more predictions into the high-overlap region.
}
\label{fig:iou_threshold_distribution}
\end{figure}

\subsection{Threshold-Sensitivity Analysis}

Figure~\ref{fig:iou_threshold_distribution} further examines how the improvement changes across IoU thresholds. Compared with the base model, \method{} improves Acc@0.5 by 2.98 points, Acc@0.7 by 4.25 points, Acc@0.9 by 7.75 points, and Acc@0.95 by 12.84 points. The IoU distribution shows the same trend: predictions below 0.5 IoU decrease from 11.4\% to 8.4\%, while predictions above 0.95 IoU increase from 35.7\% to 48.5\%. These results suggest that the privileged box-marked teacher and IoU-aware token weighting improve the overall overlap distribution, moving more predictions from low- and medium-overlap regions into high-overlap regions.

\section{Conclusion}
\label{sec:conclusion}

This paper presents \method{}, an IoU-aware privileged distillation method that uses ground-truth boxes as coordinate targets and visual guidance for a frozen teacher. Training combines supervised fine-tuning with geometry-aware distillation, while inference uses only the student with the standard image--text input. Experiments and matched ablations show that box annotations can improve grounding through teacher guidance beyond coordinate supervision alone.

\clearpage
\section*{Limitations}

\begin{itemize}
    \item The method uses ground-truth boxes as privileged information during training, so it requires grounding annotations.
    \item The gains are moderate and should be interpreted as improvements to a strong base model.
    \item The current teacher hint is a simple box overlay. More carefully designed privileged inputs may further improve the tradeoff between region recognition and coordinate precision.
\end{itemize}

\section*{Ethical Considerations}

This work studies visual grounding on standard referring-expression benchmarks. Accurate grounding can support assistive perception, human-computer interaction, and fine-grained visual understanding. However, localization methods may also be misused in surveillance, tracking, or privacy-sensitive monitoring scenarios. We do not intend the method to be used for identifying, tracking, or profiling individuals without consent. Any deployment should follow applicable privacy regulations, obtain appropriate consent, and include safeguards against harmful or unauthorized use.

\bibliography{custom}

\appendix

\begin{table*}[t]
\centering
\scriptsize
\setlength{\tabcolsep}{4pt}
\renewcommand{\arraystretch}{1.05}
\begin{tabular*}{\textwidth}{@{\extracolsep{\fill}}lrrrrrrr@{}}
\toprule
Size & \(n\) & Base Acc@0.5 & \method{} Acc@0.5 & \(\Delta_{.5}\) & Base Acc@0.7 & \method{} Acc@0.7 & \(\Delta_{.7}\) \\
\midrule
Small (\(<5\%\)) & 375 & 75.73 & 81.87 & +6.13 & 68.80 & 73.33 & +4.53 \\
Medium (\(5\%-10\%\)) & 7,953 & 84.32 & 88.67 & +4.35 & 77.10 & 82.56 & +5.46 \\
Large (\(>10\%\)) & 22,641 & 90.28 & 92.73 & +2.45 & 84.64 & 88.46 & +3.82 \\
\bottomrule
\end{tabular*}
\caption{
Source values for the object-size performance figure. Base Acc@0.5 and \method{} Acc@0.5 denote Acc@0.5 for the base model and \method{}, respectively; Base Acc@0.7 and \method{} Acc@0.7 denote Acc@0.7. \(\Delta\) values are absolute improvements in percentage points.
}
\label{tab:app_object_size_performance_source}
\end{table*}

\section{Implementation Details}
\label{appendix:implementation}

\subsection{Training Configuration}

The main \method{} setting uses Qwen3-VL-4B~\cite{qwen3vl} as the backbone, 300k grounding examples, and 3 training epochs. Training is implemented with \texttt{ms-swift}~\cite{ms}. The \texttt{--model} and \texttt{--teacher\_model} arguments are initialized from the same base checkpoint. The student receives the original image and the original referring-expression prompt. The teacher receives the box-marked image and the privileged text hint.

We use full-parameter tuning, bfloat16 precision, learning rate \(2 \times 10^{-6}\), SFT coefficient \(\alpha=1.0\), distillation temperature \(1.0\), maximum sequence length 20,000, maximum completion length 128, warmup ratio 0.05, and FlashAttention. The Qwen3-VL-2B backbone and reduced-data settings are used only for scaling ablations.

\subsection{Frozen Privileged Teacher}

The teacher has the same architecture as the student and is initialized from the same base checkpoint. During training, the teacher parameters are kept frozen. The teacher distribution is computed under the privileged input, detached from the computation graph, and used as a stop-gradient target in the distillation loss. Gradients are propagated only through the student distribution. No exponential moving average teacher is used. The teacher is used only during training and is removed at inference time.

This implementation separates the source of privileged information from the trainable student. The only difference between the student and teacher inputs is that the teacher receives a box-marked image and a short privileged hint, while the student receives the original image and the original prompt. At inference time, only the student-side input format is used.

\subsection{\method{} Hyperparameters}

The main \method{} run enables IoU-aware token weighting and uses normalized 1000-scale coordinates. The sample-level factor uses the exponential IoU form with \(\tau_r=0.5\). The coordinate-level factor uses softmax weighting with \(\tau_c=1.0\). The digit-position factor uses \(\lambda_d=0.5\). The teacher--student preference gate uses a sigmoid with \(\beta=3.0\). The teacher-confidence factor uses the entropy-based exponential form with \(\tau_e=1.0\). Token weights are detached and divided by their mean across all valid response tokens in the minibatch, with a denominator floor of \(10^{-6}\). The weighted distillation terms are averaged over the same tokens.

\subsection{Prompt and Output Format}

The same prompt format is used across training variants and evaluation. Given a referring expression \texttt{<expr>}, the student input text is formatted as:

\begin{quote}
\texttt{Please provide the bounding box coordinate of the region this sentence describes: <expr>.}
\end{quote}

The placeholder \texttt{<expr>} is replaced by the referring expression from the dataset. The target response is a four-coordinate box in the normalized coordinate system used by the dataset solution field.

For the teacher branch during training, the privileged hint is appended to the original prompt:

\begin{quote}
\texttt{The answer is located within the green rectangle.}
\end{quote}

The student and all inference-time evaluations use only the original prompt.

\section{Evaluation Protocol and Parsing Rules}
\label{appendix:evaluation_protocol}

\subsection{Unified Evaluation Protocol}

All comparison models are evaluated under the same prompt, coordinate parser, normalization rule, box canonicalization rule, and metric computation script. For models that directly output boxes, their predictions are converted into the same normalized coordinate space before metric computation. This avoids comparing results produced by different prompting, parsing, or coordinate-normalization rules.

We report mean IoU, Acc@0.5, and Acc@0.7 in the main evaluation. The paper focuses on region-level grounding rather than using high-precision boundary metrics as the central success criterion.

\subsection{Coordinate Parsing}

All predicted boxes are parsed into the normalized coordinate space before evaluation. The parser first extracts four coordinate fields from the model response. The reconstructed coordinates are canonicalized into valid box corners, clipped to the normalized coordinate range, and compared with the ground-truth box using IoU.

For malformed or incomplete responses, the parser applies the same fallback rule for all models. If four valid coordinates cannot be recovered, the prediction is treated as invalid for metric computation under the same evaluation script.

\subsection{Token-to-Coordinate Mapping}

The IoU-aware token weighting requires mapping response tokens to the geometric structure of the coordinate output. For a valid parsed response, each coordinate digit token is assigned a coordinate identity
\[
    m(t) \in \{1,2,3,4\},
\]
corresponding to \(x_1\), \(y_1\), \(x_2\), and \(y_2\). Each coordinate digit token is also assigned a digit position \(\rho(t)\), defined by decimal significance. More significant digits receive larger \(\rho(t)\). For example, in a normalized integer coordinate, the hundreds digit has a larger \(\rho(t)\) than the tens digit, and the tens digit has a larger \(\rho(t)\) than the ones digit.

Tokens that do not represent coordinate digits, including brackets, commas, spaces, separators, and punctuation tokens, are treated as non-coordinate tokens. They are assigned neutral geometry weights for the coordinate-level and digit-position factors. If the response is malformed, incomplete, or cannot be converted into four valid coordinates, the geometry-dependent factors fall back to neutral values. This keeps the distillation loss well defined for invalid responses.

\begin{table*}[t]
\centering
\scriptsize
\setlength{\tabcolsep}{4pt}
\renewcommand{\arraystretch}{1.05}
\begin{tabular*}{\textwidth}{@{\extracolsep{\fill}}lrrrrrrr@{}}
\toprule
Model & \(<0.5\) & 0.5--0.6 & 0.6--0.7 & 0.7--0.8 & 0.8--0.9 & 0.9--0.95 & \(\geq0.95\) \\
\midrule
Base & 11.42 & 2.28 & 3.78 & 6.03 & 15.25 & 25.54 & 35.70 \\
\method{} & 8.44 & 2.05 & 2.74 & 5.21 & 12.57 & 20.45 & 48.54 \\
\bottomrule
\end{tabular*}
\caption{
Predicted-IoU distribution used in Figure~\ref{fig:iou_threshold_distribution}. Bins are defined by the same IoU thresholds used in Table~\ref{tab:app_iou_threshold_accuracy}.
}
\label{tab:app_iou_distribution}
\end{table*}

\section{Additional Analysis of Privileged Hints}
\label{appendix:privileged_hints}

\begin{figure}[t]
\centering
\includegraphics[width=\linewidth]{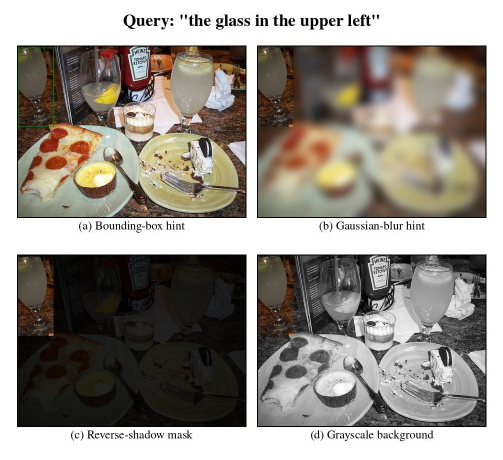}
\caption{
Comparison of different privileged visual hints. The query is ``the glass in the upper left.'' A bounding-box hint marks the target while preserving the original scene context. Other hints, such as Gaussian blur, reverse-shadow masking, or grayscale background, change the visual distribution and may remove semantic or spatial cues needed for grounding.
}
\label{fig:hint_comparison}
\end{figure}

Figure~\ref{fig:hint_comparison} explains why \method{} uses a bounding-box overlay as the privileged visual hint. The hint should identify the target region while preserving the visual context needed by the referring expression. For example, the query ``the glass in the upper left'' depends on surrounding objects and relative spatial layout. Gaussian blur, reverse-shadow masking, or grayscale background may make the target easier to isolate, but they also change the semantic and spatial structure of the image. The teacher may then rely on evidence that is not available to the student at inference time, increasing the conditional mismatch between teacher and student. A box overlay is therefore a conservative privileged hint: it provides explicit localization guidance while keeping the teacher input close to the original image.

\section{Detailed Main Results}
\label{appendix:main_setting_results}

Table~\ref{tab:appendix_main_setting_results} reports the detailed split-level comparison between the Qwen3-VL-4B base model and the main \method{} setting. The overall results are computed by pooling all examples from the five evaluation splits rather than averaging split-level scores. All values are reported as percentages.

\begin{table*}[!hbtp]
\centering
\setlength{\tabcolsep}{3.2pt}
\renewcommand{\arraystretch}{1.05}
\begin{tabular*}{\textwidth}{@{\extracolsep{\fill}}lrrrrrrrrr@{}}
\toprule
\multirow{2}{*}{Split}
& \multicolumn{3}{c}{mIoU}
& \multicolumn{3}{c}{Acc@0.5}
& \multicolumn{3}{c}{Acc@0.7} \\
\cmidrule(lr){2-4}
\cmidrule(lr){5-7}
\cmidrule(lr){8-10}
& Base & \method{} & \(\Delta\)
& Base & \method{} & \(\Delta\)
& Base & \method{} & \(\Delta\) \\
\midrule
Overall
& 81.74 & 85.78 & +4.03
& 88.58 & 91.56 & +2.98
& 82.51 & 86.76 & +4.25 \\
RefCOCO testA
& 85.90 & 88.45 & +2.55
& 93.25 & 95.19 & +1.94
& 88.56 & 91.44 & +2.88 \\
RefCOCO testB
& 81.45 & 84.20 & +2.75
& 88.85 & 90.95 & +2.10
& 81.33 & 84.14 & +2.81 \\
RefCOCOg test
& 81.85 & 87.23 & +5.38
& 88.31 & 91.45 & +3.13
& 82.18 & 87.34 & +5.16 \\
RefCOCO+ testA
& 83.79 & 87.14 & +3.34
& 90.85 & 93.59 & +2.74
& 85.98 & 89.91 & +3.93 \\
RefCOCO+ testB
& 74.64 & 79.88 & +5.24
& 80.73 & 85.80 & +5.07
& 73.33 & 79.28 & +5.95 \\
\bottomrule
\end{tabular*}
\caption{
Detailed comparison between the Qwen3-VL-4B base model and the main \method{} setting. Overall results are computed on the pooled five-split evaluation set. \(\Delta\) denotes the absolute improvement over the base model in percentage points.
}
\label{tab:appendix_main_setting_results}
\end{table*}

The main \method{} setting consistently improves over the Qwen3-VL-4B base model across all five evaluation splits. The overall gains are +4.03 mIoU, +2.98 Acc@0.5, and +4.25 Acc@0.7. The improvements are especially clear on RefCOCOg test and RefCOCO+ testB, suggesting that privileged teacher guidance is beneficial for more descriptive expressions and object- or attribute-focused grounding cases.

\section{Evaluation on Ref-L4}
\label{appendix:ref_l4}

We evaluate \method{} on Ref-L4 to test whether its grounding gains extend to more diverse objects and more detailed referring expressions. Ref-L4~\cite{chen2025revisiting} covers a broader set of object categories and uses longer expressions than the RefCOCO benchmarks. It therefore provides a more demanding setting for identifying and localizing the referred object. We use the official test split and compare with Qwen3-VL-4B-Instruct, Hi-Token~\cite{zhu2026hitoken}, and VLM-R1~\cite{vlm-r1}.

As shown in Table~\ref{tab:ref_l4_results}, \method{} improves mIoU by 4.33 percentage points and Acc@0.5 by 2.40 percentage points over the base model. It also achieves the highest scores among the four evaluated models on both metrics. These results extend our findings to a benchmark with richer descriptions and broader category coverage.

\begin{table}[!htbp]
\centering
\setlength{\tabcolsep}{5pt}
\renewcommand{\arraystretch}{1.05}
\begin{tabular*}{\columnwidth}{@{\extracolsep{\fill}}lrr@{}}
\toprule
Model & mIoU (\%) & Acc@0.5 (\%) \\
\midrule
Base & 76.43 & 84.70 \\
Hi-Token & 62.71 & 65.53 \\
VLM-R1 & 70.10 & 76.63 \\
\method{} & \textbf{80.76} & \textbf{87.10} \\
\bottomrule
\end{tabular*}
\caption{Visual grounding results on the official Ref-L4 test split. Base denotes Qwen3-VL-4B-Instruct. Both metrics are reported as percentages.}
\label{tab:ref_l4_results}
\end{table}

\section{Ablation Configurations}
\label{appendix:ablation_configurations}

The ablations in Table~\ref{tab:component_ablations_summary} use the same main 4B, 300k, 3-epoch setting unless otherwise specified. The variants differ only in training switches.

Variant A uses SFT only. It disables the teacher branch and disables IoU-aware token weighting. Variant B is the non-privileged self-distillation baseline. It keeps the teacher branch but feeds the original image to both teacher and student, so the teacher does not receive a box-marked image. Variant C removes the SFT anchor by setting \(\alpha=0\), while keeping the privileged teacher and IoU-aware weighting. Variant D keeps SFT and the privileged teacher but disables IoU-aware token weighting. Variant E keeps IoU-aware token weighting but disables the sample-level IoU factor \(r(u)\). Variant F keeps all components but uses a weaker sample-level temperature \(\tau_r=1.0\). The full model uses SFT, the frozen privileged teacher, IoU-aware token weighting, the sample-level IoU factor, and \(\tau_r=0.5\).

These configurations are designed to separate the effects of direct coordinate supervision, non-privileged self-distillation, privileged box-marked teacher input, IoU-aware token weighting, and sample-level IoU weighting.

\section{Additional Quantitative Analyses}
\label{appendix:figure_tables}

This section provides the numerical values behind the analysis figures in the main paper and gives additional interpretation of these results. The goal is not only to document the plotted values, but also to clarify what each analysis measures and how it supports the design choices of \method{}.

\subsection{Component Ablation Analysis}
\label{app:component_ablation_values}

Table~\ref{tab:app_component_ablations_source} reports the full component ablation results used to generate Figure~\ref{fig:component_ablations_single}. All rows use the main 4B, 300k, 3-epoch setting unless otherwise specified. The variants are designed to isolate five factors: supervised fine-tuning, teacher distillation, privileged box-marked teacher input, IoU-aware token weighting, and the sample-level IoU factor \(r(u)\).

\begin{table*}[t]
\centering
\scriptsize
\setlength{\tabcolsep}{3.2pt}
\renewcommand{\arraystretch}{1.05}
\begin{tabular*}{\textwidth}{@{\extracolsep{\fill}}llccccccrrrrr@{}}
\toprule
Variant & Setting & SFT & Teacher & Box & Weight & \(r(u)\) & \(\tau_r\)
& mIoU & Acc@0.5 & \(\Delta_{.5}\) & Acc@0.7 & \(\Delta_{.7}\) \\
\midrule
Base & Base model
& \(\times\) & \(\times\) & \(\times\) & \(\times\) & \(\times\) & --
& 0.8174 & 88.58 & -- & 82.51 & -- \\
A & SFT only
& \(\checkmark\) & \(\times\) & \(\times\) & \(\times\) & \(\times\) & --
& 0.8470 & 90.62 & +2.04 & 85.32 & +2.81 \\
B & Original teacher
& \(\checkmark\) & \(\checkmark\) & \(\times\) & \(\times\) & \(\times\) & --
& 0.8492 & 89.80 & +1.22 & 85.63 & +3.12 \\
C & No SFT anchor
& \(\times\) & \(\checkmark\) & \(\checkmark\) & \(\checkmark\) & \(\checkmark\) & 0.5
& 0.8335 & 89.42 & +0.84 & 83.76 & +1.25 \\
D & Box teacher
& \(\checkmark\) & \(\checkmark\) & \(\checkmark\) & \(\times\) & \(\times\) & --
& 0.8543 & 91.23 & +2.65 & 86.28 & +3.77 \\
E & Token weighting
& \(\checkmark\) & \(\checkmark\) & \(\checkmark\) & \(\checkmark\) & \(\times\) & --
& 0.8565 & 91.45 & +2.87 & 86.55 & +4.04 \\
F & Sample-level factor
& \(\checkmark\) & \(\checkmark\) & \(\checkmark\) & \(\checkmark\) & \(\checkmark\) & 1.0
& 0.8570 & 91.50 & +2.92 & 86.63 & +4.12 \\
Full & Full \method{}
& \(\checkmark\) & \(\checkmark\) & \(\checkmark\) & \(\checkmark\) & \(\checkmark\) & 0.5
& \textbf{0.8578} & \textbf{91.56} & \textbf{+2.98} & \textbf{86.76} & \textbf{+4.25} \\
\bottomrule
\end{tabular*}
\caption{
Full component ablation results. SFT denotes supervised fine-tuning, Teacher denotes distillation from a teacher branch, Box denotes privileged box-marked teacher input, Weight denotes IoU-aware token weighting, and \(r(u)\) denotes the sample-level IoU factor. \(\Delta_{.5}\) and \(\Delta_{.7}\) are absolute Acc@0.5 and Acc@0.7 improvements over the 4B base model.
}
\label{tab:app_component_ablations_source}
\end{table*}

The ablation results show that SFT is the strongest single component. Variant A improves Acc@0.7 from 82.51 to 85.32, giving a +2.81 point gain over the base model. This indicates that direct coordinate supervision is important for adapting the model to the output format and grounding distribution.

The comparison between variants B and D isolates the effect of privileged visual input. Variant B uses a teacher branch without a box-marked image, while variant D adds the box-marked teacher input. Moving from B to D improves mIoU from 0.8492 to 0.8543 and Acc@0.7 from 85.63 to 86.28. This suggests that the gain is not merely from adding a teacher branch; the teacher-side box mark provides additional training-time visual guidance.

Variant C removes the SFT anchor while keeping the privileged teacher and IoU-aware weighting. It still improves over the base model, but it is clearly weaker than SFT-only training and the full model. This supports the choice of using privileged distillation as a complement to direct coordinate supervision rather than as a replacement for it.

The remaining rows show the effect of the weighting design. Adding token weighting improves variant D from 86.28 to 86.55 Acc@0.7. Adding the sample-level IoU factor further improves performance, and the full setting with \(\tau_r=0.5\) gives the best result. The gains from the last weighting components are smaller than the gain from SFT or privileged box input, but they are consistent across mIoU, Acc@0.5, and Acc@0.7.

\subsection{Scaling Ablation Analysis}
\label{app:scaling_ablation_values}

Table~\ref{tab:app_scaling_ablations_source} reports the metrics used in Figure~\ref{fig:scaling_ablations}. Deltas are computed against the corresponding same-size base model. These results analyze whether the improvement depends on a particular training scale or remains visible under reduced settings.

\begin{table*}[t]
\centering
\scriptsize
\setlength{\tabcolsep}{4pt}
\renewcommand{\arraystretch}{1.05}
\begin{tabular*}{\textwidth}{@{\extracolsep{\fill}}lccrrrrr@{}}
\toprule
Setting & Data & Epochs & mIoU & Acc@0.5 & \(\Delta_{.5}\) & Acc@0.7 & \(\Delta_{.7}\) \\
\midrule
Base 2B & -- & -- & 0.7931 & 85.64 & -- & 79.76 & -- \\
2B & 30k & 1 & 0.7993 & 86.64 & +0.99 & 80.63 & +0.87 \\
\midrule
Base 4B & -- & -- & 0.8174 & 88.58 & -- & 82.51 & -- \\
4B & 30k & 1 & 0.8234 & 89.26 & +0.68 & 83.20 & +0.68 \\
4B & 30k & 3 & 0.8304 & 90.05 & +1.47 & 84.13 & +1.62 \\
4B & 30k & 5 & 0.8310 & 90.16 & +1.58 & 84.16 & +1.65 \\
4B & 80k & 1 & 0.8288 & 89.94 & +1.36 & 83.84 & +1.33 \\
4B & 300k & 1 & 0.8393 & 90.94 & +2.36 & 85.38 & +2.87 \\
Full \method{} 4B & 300k & 3 & \textbf{0.8578} & \textbf{91.56} & \textbf{+2.98} & \textbf{86.76} & \textbf{+4.25} \\
\bottomrule
\end{tabular*}
\caption{
Scaling ablation results. The table reports the metrics behind the data-scale, epoch-budget, and final-configuration panels in Figure~\ref{fig:scaling_ablations}. \(\Delta_{.5}\) and \(\Delta_{.7}\) denote absolute Acc@0.5 and Acc@0.7 improvements over the same-size base model.
}
\label{tab:app_scaling_ablations_source}
\end{table*}

The reduced 2B setting improves over the 2B base model by +0.87 Acc@0.7, showing that the training strategy is still beneficial with a smaller backbone. For the 4B backbone, increasing the data size under a fixed one-epoch budget gives larger gains: +0.68 Acc@0.7 with 30k examples, +1.33 with 80k examples, and +2.87 with 300k examples. This indicates that the method benefits from more grounding data.

The epoch-budget comparison shows a different trend. With 30k examples, increasing training from 1 to 3 epochs improves Acc@0.7 from 83.20 to 84.13, but increasing further to 5 epochs gives only a marginal improvement to 84.16. This suggests that, under limited data, additional epochs quickly saturate. The strongest setting is therefore obtained by combining larger data scale with sufficient optimization budget: the full 4B, 300k, 3-epoch setting reaches 86.76 Acc@0.7, corresponding to a +4.25 point gain over the 4B base model.

\subsection{Object-Size Analysis}
\label{app:object_size_values}

Table~\ref{tab:app_object_size_performance_source} reports the object-size breakdown used in Figure~\ref{fig:object_size_performance}. Objects are grouped by the ground-truth box area in the normalized coordinate space. This analysis examines whether the method only improves easy large-object cases or also helps smaller targets.

The method improves all object-size groups. The gains are not limited to large objects: small objects improve by +6.13 points at Acc@0.5 and +4.53 points at Acc@0.7, while medium objects improve by +4.35 and +5.46 points. Large objects also improve, although the gain is smaller at Acc@0.5 because the base model is already stronger on this group. The small-object group contains fewer examples than the medium and large groups, so the exact magnitude should be interpreted with this sample size in mind. Still, the consistent gains across all three groups suggest that the method improves grounding beyond only the easiest large-object cases.

\subsection{IoU Threshold Accuracy Analysis}
\label{app:iou_threshold_values}

Table~\ref{tab:app_iou_threshold_accuracy} reports thresholded grounding accuracy from Acc@0.5 to Acc@0.95. The evaluation aggregates the five held-out splits, with \(n=30{,}969\) examples per model. This analysis measures how the improvement changes as the IoU threshold becomes stricter.

\begin{table}[!htbp]
\centering
\scriptsize
\setlength{\tabcolsep}{2pt}
\renewcommand{\arraystretch}{1.05}
\begin{tabular*}{\columnwidth}{@{\extracolsep{\fill}}lrrrrrr@{}}
\toprule
Model & Acc@0.5 & Acc@0.6 & Acc@0.7 & Acc@0.8 & Acc@0.9 & Acc@0.95 \\
\midrule
Base & 88.58 & 86.29 & 82.51 & 76.49 & 61.24 & 35.70 \\
\method{} & 91.56 & 89.51 & 86.76 & 81.56 & 68.99 & 48.54 \\
\(\Delta\) & +2.98 & +3.21 & +4.25 & +5.07 & +7.75 & +12.84 \\
\bottomrule
\end{tabular*}
\caption{
Thresholded grounding accuracy. P@\(t\) is the percentage of examples whose predicted box reaches IoU threshold \(t\).
}
\label{tab:app_iou_threshold_accuracy}
\end{table}

The gain becomes larger under stricter thresholds. The improvement is +2.98 points at Acc@0.5, +4.25 points at Acc@0.7, +7.75 points at Acc@0.9, and +12.84 points at Acc@0.95. This does not change the main focus of the paper, which is region-level grounding, but it shows that the improvement is also reflected in the higher-overlap part of the IoU spectrum. In other words, the method does not merely convert very poor predictions into loosely correct ones; it also shifts many already-correct predictions toward higher overlap.

\subsection{IoU Distribution Analysis}
\label{app:iou_distribution_values}

Table~\ref{tab:app_iou_distribution} reports the binned IoU distribution used in Figure~\ref{fig:iou_threshold_distribution}. Each row sums to 100\%, up to rounding. This distribution provides a complementary view to the thresholded accuracy table.

The fraction of predictions below 0.5 IoU decreases from 11.42\% to 8.44\%. Predictions with IoU at least 0.95 increase from 35.70\% to 48.54\%, while several intermediate bins contain a smaller fraction of predictions. Together with the threshold results, this distribution shows that the complete method produces more boxes with high geometric overlap than Base.

\subsection{Token-Weighting Analysis}
\label{app:token_weighting_values}

Table~\ref{tab:app_token_weighting_source} reports the final token weights and coordinate-level means used in Figure~\ref{fig:token_weighting}. The example response is \([180, 220, 600, 660]\), and each coordinate is decomposed into hundreds, tens, and ones digits. This example illustrates how the final weights vary across both coordinates and digit positions.

\begin{table}[!htbp]
\centering
\setlength{\tabcolsep}{4pt}
\renewcommand{\arraystretch}{1.05}
\begin{tabular*}{\columnwidth}{@{\extracolsep{\fill}}lcrrrr@{}}
\toprule
Coordinate & Digits & H & T & O & Mean \\
\midrule
\(x_1\) & 180 & 0.412 & 0.328 & 0.195 & 0.312 \\
\(y_1\) & 220 & 1.480 & 1.078 & 0.517 & 1.025 \\
\(x_2\) & 600 & 1.223 & 0.848 & 0.452 & 0.841 \\
\(y_2\) & 660 & 2.687 & 1.891 & 0.889 & 1.822 \\
\bottomrule
\end{tabular*}
\caption{
Token-weighting example. H, T, and O denote the hundreds, tens, and ones digit positions. The mean column is the coordinate-level average final token weight shown in Figure~\ref{fig:token_weighting}.
}
\label{tab:app_token_weighting_source}
\end{table}

The weights are not uniform across the coordinate string. Within each coordinate, the hundreds digit receives a larger weight than the tens digit, and the tens digit receives a larger weight than the ones digit. This follows the digit-position design, where more significant digits have a larger effect on the decoded coordinate. Across coordinates, the mean weights also differ. In this example, \(y_2\) receives the largest coordinate-level mean weight, while \(x_1\) receives the smallest. This reflects the combined effect of coordinate-level error, digit position, teacher--student preference, and teacher confidence. The example therefore illustrates that the distillation loss is adapted to the geometric structure of the coordinate output rather than applied uniformly to all response tokens.

\section{Artifact Use and Documentation}
\label{appendix:artifacts}

\subsection{Artifact Use and Licenses}

This work uses publicly available research artifacts, including Qwen3-VL models, RefCOCO grounding benchmarks, and open-source training or inference software. These artifacts are used for research on visual grounding, which is consistent with their intended research use.

The original datasets are not redistributed. Any released code, trained checkpoints, or derived artifacts will be intended for research use and should follow the licenses and terms of the underlying datasets, models, and software frameworks.

\subsection{Artifact Documentation}

This work uses existing research artifacts for visual grounding. The evaluated datasets are standard referring-expression grounding benchmarks built on natural images and English referring expressions. They cover object localization from language descriptions in general visual scenes. No new dataset or human annotation is introduced in this work.

The main model artifact used in the experiments is Qwen3-VL-4B~\cite{qwen3vl}. Additional scaling ablations use Qwen3-VL-2B~\cite{qwen3vl}. The software artifacts include \texttt{ms-swift}~\cite{ms} for training and vLLM for rollout or inference support. These artifacts are used for research on coordinate-generating visual grounding. The original creators of the datasets, models, and software tools are cited in the main paper.

\subsection{AI Assistants in Research and Writing}

During the preparation of this paper, AI assistants were used to support language polishing, LaTeX editing, and code debugging. All experiments, analyses, and interpretations were conducted, checked, and approved by the authors. The authors are fully responsible for the accuracy and integrity of the paper.

\end{document}